\documentclass{article}
\usepackage{ijcai19}

\usepackage{url,graphicx,tabularx,array, datetime, booktabs, mathrsfs, color, tablefootnote, tabu} 
\usepackage{times}
\usepackage{soul}
\usepackage{url}
\usepackage[hidelinks]{hyperref}
\usepackage[utf8]{inputenc}
\usepackage[small]{caption}
\usepackage{graphicx}
\usepackage{amsmath}
\usepackage{booktabs}	
\usepackage{algorithm}
\usepackage{algorithmic}
\usepackage{svg} 
\usepackage{amsmath, bm, amsfonts, cases, algorithmic, enumitem}

\newcommand{\meta}{Meta-NLG}
\newcommand{\mtl}{MTL-NLG}
\newcommand{\scratch}{Scratch-NLG}
\newcommand{\zero}{Zero-NLG}
\newcommand{\supervise}{Supervised-NLG}
\newcommand*{\red}{\textcolor{red}}
\newcommand*{\blue}{\textcolor{blue}}

\newcommand*{\orange}{\textcolor{orange}}

\title{Meta-Learning for Low-resource Natural Language Generation in Task-oriented Dialogue Systems}
\author{
Fei Mi$^1$\footnote{Contact author; this work was done when Fei Mi was a visiting scholar at Tsinghua University.}\and
Minlie Huang$^2$\and
Jiyong Zhang$^{3}$\And
Boi Faltings$^1$\\
\affiliations
$^1$École polytechnique fédérale de Lausanne (EPFL)\\
$^2$Tsinghua University\\
$^3$Hangzhou Dianzi University\\
\emails
fei.mi@epfl.ch,
aihuang@tsinghua.edu.cn,
jzhang@hdu.edu.cn,
boi.faltings@epfl.ch
}

\begin{document}
	
	\maketitle
	
	\begin{abstract}
		
		Natural language generation (NLG) is an essential component of task-oriented dialogue systems.
		Despite the recent success of neural approaches for NLG, they are typically developed for particular domains with rich annotated training examples.
		In this paper, we study NLG in a low-resource setting to generate sentences in new scenarios with handful training examples.
		We formulate the problem from a meta-learning perspective, and propose a generalized optimization-based approach~(\meta) based on the well-recognized model-agnostic meta-learning~(MAML) algorithm.
		\meta\ defines a set of meta tasks, and directly incorporates the objective of adapting to new low-resource NLG tasks into the meta-learning optimization process.
		Extensive experiments are conducted on a large multi-domain dataset (MultiWoz) with diverse linguistic variations. 
		We show that \meta\ significantly outperforms other training procedures in various low-resource configurations. 
		We analyze the results, and demonstrate that \meta\ adapts extremely \textit{fast and well }to low-resource situations.
		
	\end{abstract}
	\vspace{-0.15in}
	\section{Introduction}
	
	As an essential part of a task-oriented dialogue system~\cite{wen2016network}, the task of natural language generation (NLG) is to produce a natural language utterance containing the desired information given a \textit{semantic representation} consisting of dialogue act types with a set of slot-value pairs.
	Conventional methods using hand-crafted rules often generates monotonic utterances and it requires substantial amount of human engineering work.
	Recently, various neural approaches~\cite{wen2015semantically,tran2017natural,tseng2018variational} have been proposed to generate accurate, natural and diverse utterances.
	However, these methods are typically developed for particular domains. Moreover, they are often data-intensive to train. The high annotation cost prevents developers to build their own NLG component from scratch. 
	Therefore, it is extremely useful to train a NLG model that can be generalized to other NLG domains or tasks with a reasonable amount of annotated data. This is referred to \textit{low-resource} NLG task in this paper.
	
	Recently, some methods have been proposed for low-resource NLG tasks.
	Apart from the simple data augmentation trick~\cite{wen2016multi}, specialized model architectures, including conditional variational auto-encoders (CVAEs, \cite{tseng2018variational,tran2018adversarial,tran2018dual}) and adversarial domain adaptation critics~\cite{tran2018adversarial}, have been proposed to learn domain-invariant representations.
	Although promising results were reported, we found that datasets used by these methods are simple which tend to enumerate many slots and values in an utterance without much linguistic variations.
	As a consequence, over-fitting the slots and values in the low-resource target domain could even outperform those versions trained with rich source domain examples~\cite{tran2018dual}.
	Fortunately, there is a new large-scale dialog dataset~(MultiWoz, \cite{budzianowski2018multiwoz}) that contains a great variety of domains and linguistic patterns that allows us to conduct extensive and meaningful experimental analysis for low-resource NLG tasks. 
	
	In this paper, instead of casting the problem as model-based approaches, we propose a generalized optimization-based meta-learning approach to directly enhance the optimization procedure for the low-resource NLG task.
	We start by arguing that a recently proposed model-agnostic meta-learning algorithm (MAML, \cite{finn2017model}) is a nice fit to the low-resource NLG task. Then, we proposed a generalized NLG algorithm called \meta\ based on MAML by viewing languages in different domains or dialog act types as separate \textit{Meta NLG tasks}. Following the essence of MAML, the goal of \meta\ is to learn a better initialization of model parameters that facilitates fast adaptation to new low-resource NLG scenarios.
	As \meta\ is model-agnostic as long as the model can be optimized by gradient descent, we could apply it to any existing NLG models to optimize them in a way that adapt better and faster to new low-resource tasks. 
	
	The main contribution of this paper is two-fold:
	\begin{itemize}
		\item 	We propose a meta-learning algorithm \meta\ based on MAML for low-resource NLG tasks. Since \meta\ is model-agnostic, it is applicable to many other NLG models. To the best of our knowledge, this is the first study of applying meta-learning to NLG tasks.
		
		\item  We extensively evaluate \meta\ on the largest multi-domain dataset (MultiWoz) with various low-resource NLG scenarios.
		Results show that \meta\ significantly outperforms other optimization methods in various configurations. We further analyze the superior performance of \meta, and show that it indeed adapts much faster and better.
	\end{itemize}
	
	\section{Background}
	
	\subsection{Natural Language Generation (NLG)}
	Neural models have recently shown promising results in tackling NLG tasks for task-oriented dialog systems. 
	Conditioned on some \textit{semantic representation} called dialog act~(DA), a NLG model decodes an utterance word by word, and the probability of generating an output sentence of length $T$ is factorized as below:
	\vspace{-0.05in}
	\begin{equation}
	\small
	f_{\theta} =  P(\mathbf{Y}|\mathbf{d}; \theta) = \prod_{t=1}^{T} P(y_{t} | y_{0} ,..., y_{t-1}, \mathbf{d}; \theta)
	\end{equation}
	
	$f_{\theta}$ is the NLG model parameterized by $\theta$, and $\mathbf{d}$ is the DA of sentence $\mathbf{Y} = (y_0, y_1, ..., y_{T})$. For example, $\mathbf{d}$ is a one-hot representation of a DA ``Inform(name=The Oak Bistro, food=British)''. ``Inform'' (DA type) controls the sentence functionality, and ``name'' and ``food'' are two involved slots. A \textit{realization utterance} $\mathbf{Y}$ could be ``\textit{There is a restaurant called [The Oak Bistro] that serves [British] food.}''. Each sentence might contain multiple DA types. 
	A series of neural methods have been proposed, including HLSTM~\cite{wen2015stochastic}, SCLSTM~\cite{wen2015semantically}, Enc-Dec~\cite{wen2016toward} and RALSTM~\cite{tran2017natural}.
	
	\subsection{Low-resource NLG}
	The goal of low-resource NLG is to fine-tune a pre-trained NLG model on new NLG tasks (e.g., new domains) with a small amount of training examples.
	\cite{wen2016multi} proposed a ``data counterfeiting'' method to augment the low-resource training data in the new task without modifying the model or training procedure.
	\cite{tseng2018variational} proposed a semantically-conditioned variational autoencoder (SCVAE) learn domain-invariant representations feeding to SCLSTM. They shown that it improves SCLSTM in low-resource settings.
	\cite{tran2018dual} adopted the same idea as in \cite{tseng2018variational}. They used two conditional variational autoencoders to encode the sentence and the DA into two separate latent vectors, which are fed together to the decoder RALSTM~\cite{tran2017natural}.
	They later designed two domain adaptation critics with an adversarial training algorithm~\cite{tran2018adversarial} to learn an indistinguishable latent representation of the source and the target domain to better generalize to the target domain.
    Different from these model-based approaches, we directly tackle the optimization issue from a meta-learning perspective.

	\subsection{Meta-Learning}

	Meta-learning or learning-to-learn, 
	which can date back to some early works~\cite{naik1992meta},
	has recently attracted extensive attentions.
	A fundamental problem is ``fast adaptation to new and limited observation data''. In pursuing this problem, there are three categories of meta-learning methods:
	
	\textbf{Metric-based:} The idea is to learn a metric space and then  use it to compare low-resource testing samples to rich training samples.
	The representative works in this category include Siamese Network~\cite{koch2015siamese}, Matching Network~\cite{vinyals2016matching}, Memory-augmented Neural Network~(MANN~\cite{santoro2016meta}), Prototype Net~\cite{snell2017prototypical}, and Relation Network~\cite{sung2018learning}.
	
	\textbf{Model-based:}  The idea is to use an additional meta-learner to learn to update the original learner with a few training examples. 
	\cite{andrychowicz2016learning} developed a meta-learner based on LSTMs.  
	Hypernetwork~\cite{ha2016hypernetworks}, MetaNet~\cite{munkhdalai2017meta}, and TCML~\cite{mishra2017meta} also learn a separate set of representations for fast model adaptation.
	\cite{ravi2017optimization} proposed an LSTM-based meta-learner to learn the optimization algorithm~(gradients) used to train the original network.
	
	\textbf{Optimization-based: } The optimization algorithm itself can be designed in a way that favors fast adaption. 
	Model-agnostic meta-learning (MAML,~\cite{finn2017model,yoon2018bayesian,gu2018meta}) achieved state-of-the-art performance by directly optimizing the gradient towards a good parameter initialization for easy fine-tuning on low-resource scenarios.  It introduces no additional architectures nor parameters. Reptile~\cite{nichol2018reptile} is similar to MAML with only first-order gradient.
    In this paper, we propose a generalized meta optimization method based on MAML to directly solve the intrinsic learning issues of low-resource NLG tasks.
	

	\section{Meta-Learning for Low-resource NLG}
	
	In this section, we first describe the objective of fine-tuning a NLG model on a low-resource NLG task in Section 3.1. Then, we describe how our \meta\ algorithm encapsulates this objective into \textit{Meta NLG tasks} and into the meta optimization algorithm to learn better low-resource NLG models.
	
	\subsection{Fine-tune a NLG model}	
	Suppose $f_{\theta}$ is the base NLG model parameterized by $\theta$, and we have an initial $\theta^{\mathbf{s}}$ pre-trained with DA-utterance pairs $\mathcal{D}_\mathbf{s} = \{(\mathbf{d}_j, \mathbf{Y}_j)\}_{j\in \mathbf{s}}$ from a set $\mathbf{s}$ of high-resource source tasks. When we adapt $f_{\theta}$ to some low-resource task $t$ with DA-utterance pairs $\mathcal{D}_t = (\mathbf{d}_t, \mathbf{Y}_t)$, the fine-tuning process on $\mathcal{D}_t$ can be formulated as follows:
	
	\begin{equation}
	\small
	\begin{aligned}
	\theta^{*}  & = Adapt (\mathcal{D}_t, \theta=\theta^\mathbf{s})  = \arg \max_{\theta} 
	\mathcal{L}_{{\mathcal{D}_{t}}}(f_{\theta}) \\
	& =  \arg \max_{\theta} \sum_{(\mathbf{d}_t, \mathbf{Y}_t) \in \mathcal{D}_t} log  P(\mathbf{Y}_t|\mathbf{d}_t; \theta)
	\end{aligned}
	\label{eq:finetune}
	\end{equation}
	
	The parameter $\theta^{\mathbf{s}}$ will be used for initialization, and the model is further updated by new observations $\mathcal{D}_t$.
	The size of $\mathcal{D}_t$ in low-resource NLG tasks is very small due to the high annotation cost, therefore, a good initialization parameter $\theta^\mathbf{s}$ learned from high-resource source tasks is crucial for the adaptation performance on new low-resource NLG tasks.

	\subsection{Meta NLG Tasks}
	
	To learn a $\theta^\mathbf{s}$ that can be easily fine-tuned on new low-resource NLG tasks, the idea of our \meta\ algorithm is to repeatedly simulate auxiliary \textit{Meta NLG tasks} from $\mathcal{D}_\mathbf{s}$ to mimic the fine-tuning process in Eq.(\ref{eq:finetune}). Then, we treat each \textit{Meta NLG task} as a single meta training sample/episode, and utilize the meta optimization objective in the next section to directly learn from them.
	
	Therefore, the first step is to construct a set of auxiliary \textit{Meta NLG tasks} $(\mathcal{T}_1, ..., \mathcal{T}_K)$ to simulate the low-resource fine-tuning process. We construct a \textit{Meta NLG task} $\mathcal{T}_i$ by:
	\vspace{-0.05in}
	\begin{equation}
	\small
	\mathcal{T}_i = (\mathcal{D}_{\mathcal{T}_i}, \mathcal{D}_{\mathcal{T}_i}^{'})
	\label{eq:task}
	\vspace{-0.02in}
	\end{equation}
    $\mathcal{D}_{\mathcal{T}_i}$ and $ \mathcal{D}_{\mathcal{T}_i}^{'}$ of each $\mathcal{T}_i$ are two independent subsets of DA-utterance pairs from high-resource source data $\mathcal{D}_\mathbf{s}$. 
	$\mathcal{D}_{\mathcal{T}_i}$ and $ \mathcal{D}_{\mathcal{T}_i}^{'}$ correspond to meta-train (support) and meta-test (query) sets of a typical meta-learning or few-shot learning setup, and $\mathcal{T}_i$ is often referred to as a training episode. 
	This meta setup with both $\mathcal{D}_{\mathcal{T}_i}$ and $\mathcal{D}_{\mathcal{T}_i}^{'}$ in one \textit{Meta NLG task} allows our \meta\ algorithm to directly learn from different {\textit{Meta NLG tasks}}. The usage of them will be elaborated later. 
	\textit{Meta NLG tasks} are constructed with two additional principles:
	
	\textbf{Task Generalization}: To generalize to new NLG tasks, \textit{Meta NLG tasks} follow the same modality as the target task. 
	For example, if our target task is to adapt to DA-utterance pairs in a new domain, then DA-utterance pairs in each $\mathcal{T}_i$ are sampled from the same source domain. We also consider adapting to new DA types in later experiments. In this case, DA-utterance pairs in each $\mathcal{T}_i$ have the same DA type. This setting merges the goal of task generalization.
	
	\textbf{Low-resource Adaptation:} To simulate the process of adapting to a low-resource NLG task, the sizes of both subsets $\mathcal{D}_{\mathcal{T}_i}$ and $ \mathcal{D}_{\mathcal{T}_i}^{'}$, especially $\mathcal{D}_{\mathcal{T}_i}$, are set small. Therefore, when the model is updated on $\mathcal{D}_{\mathcal{T}_i}$ as a part of the later meta-learning steps, it only sees a small amount of samples in that task. This setup embeds the goal of low-resource adaptation.
	
	\subsection{Meta Training Objective}
	
	With the \emph{Meta NLG tasks} defined above, we formulate the meta-learning objective of \meta\ as below:
	
	\begin{equation}
	\small
	\begin{aligned}
	\theta^{Meta} &  = MetaLearn (\mathcal{T}_1, ..., \mathcal{T}_K) \\
	&  = \arg \max_{\theta} \mathbb{E}_{i}
	\mathbb{E}_{{\mathcal{D}_{\mathcal{T}_i}}, \mathcal{D}_{\mathcal{T}_i}^{'}} \mathcal{L}_{\mathcal{D}_{\mathcal{T}_i}^{'}}( f_{\theta_i^{'}} )\\
	\end{aligned}
	\label{eq:meta_ojb}
	\vspace{-0.02in}
	\end{equation}
	
	\begin{equation}
	\small
	\theta^{'}_i = Adapt (\mathcal{D}_{\mathcal{T}_i}, \theta)   = \theta - \alpha \nabla_\theta \mathcal{L}_{\mathcal{D}_{\mathcal{T}_i}}(f_{\theta})
	\label{eq:inner}
	\end{equation}
	
	The optimization for each {Meta NLG task} $\mathcal{T}_i$ is computed on $ \mathcal{D}_{\mathcal{T}_i}^{'}$ referring to $ \mathcal{D}_{\mathcal{T}_i}$.
	\textbf{Firstly}, the model parameter $\theta$ to be optimized is updated on $\mathcal{D}_{\mathcal{T}_i}$ by Eq.(\ref{eq:inner}). This step mimics the process when $f_{\theta}$ is adapted to a new low-resource NLG task $\mathcal{T}_i$ with low-resource observations $\mathcal{D}_{\mathcal{T}_i}$. 
	We need to note that Eq.(\ref{eq:inner}) is an intermediate step, and it only provides an adapted parameter~($\theta_{i}^{'}$) to our base model $f$ to be optimized in each iteration.  
	\textbf{Afterwards}, the base model parameterized by the updated parameter~($\theta_{i}^{'}$) is optimized on $\mathcal{D}_{\mathcal{T}_i}^{'}$ using the meta objective in Eq.(\ref{eq:meta_ojb}).
	This meta-learning optimization objective directly optimizes the model towards generalizing to new low-resource NLG tasks by simulating the process repeatedly with \textit{Meta NLG tasks} in Eq.(\ref{eq:meta_ojb}).
	
	The optimization of Eq.(\ref{eq:meta_ojb}) can be derived in Eq.(\ref{eq:outer}).
	It involves a standard first-order gradient $\nabla_{\theta_i^{'}} \mathcal{L}_{\mathcal{D}_{\mathcal{T}_i}^{'}} (f_{\theta_i^{'}} ) $ as well as a gradient
	through another gradient $\nabla_{\theta} {(\theta_i^{'})}$.
	Previous study~\cite{finn2017model} shows that the second term can be approximated for computation efficiency with marginal performance drop. 
	In our case, we still use the exact optimization in Eq.(\ref{eq:outer}) as we do not encounter any computation difficulties even on the largest NLG dataset so far.
	The second-order gradient is computed by a Hessian matrix $H_\theta$.
	
	\vspace{-0.15in}
	\begin{equation}
	\small
	\begin{aligned}
	\theta^{*} & = \theta - \beta \sum_{i=1}^{K}\nabla_\theta \mathcal{L}_{\mathcal{D}_{\mathcal{T}_i}^{'}} (f_{\theta_i^{'}}) \\
	& =  \theta - \beta \sum_{i=1}^{K}\nabla_{\theta_i^{'}} \mathcal{L}_{\mathcal{D}_{\mathcal{T}_i}^{'}}(f_{\theta_i^{'}}) \cdot  \nabla_{\theta}  ( {\theta_i^{'}}) \\
	& =  \theta - \beta \sum_{i=1}^{K}\nabla_{\theta_i^{'}} \mathcal{L}_{\mathcal{D}_{\mathcal{T}_i}^{'}} (f_{\theta_i^{'}}) \cdot  \nabla_{\theta} (\theta - \alpha \nabla_\theta \mathcal{L}_{\mathcal{D}_{\mathcal{T}_i}} (f_{\theta}))\\
	& = \theta - \beta \sum_{i=1}^{K} \nabla_{\theta_i^{'}} \mathcal{L}_{\mathcal{D}_{\mathcal{T}_i}^{'}} (f_{\theta_i^{'}} ) -  \alpha \nabla_{\theta_i^{'}} \mathcal{L}_{\mathcal{D}_{\mathcal{T}_i}^{'}} (f_{\theta_i^{'}} )\cdot H_\theta( \mathcal{L}_{\mathcal{D}_{\mathcal{T}_i}} (f_{\theta}))
	\end{aligned}
	\label{eq:outer}
	\vspace{-0.08in}
	\end{equation}
	\vspace{-0.25in}
    \begin{figure}[htb!]
		\centering
		\includegraphics[width=0.45\textwidth]{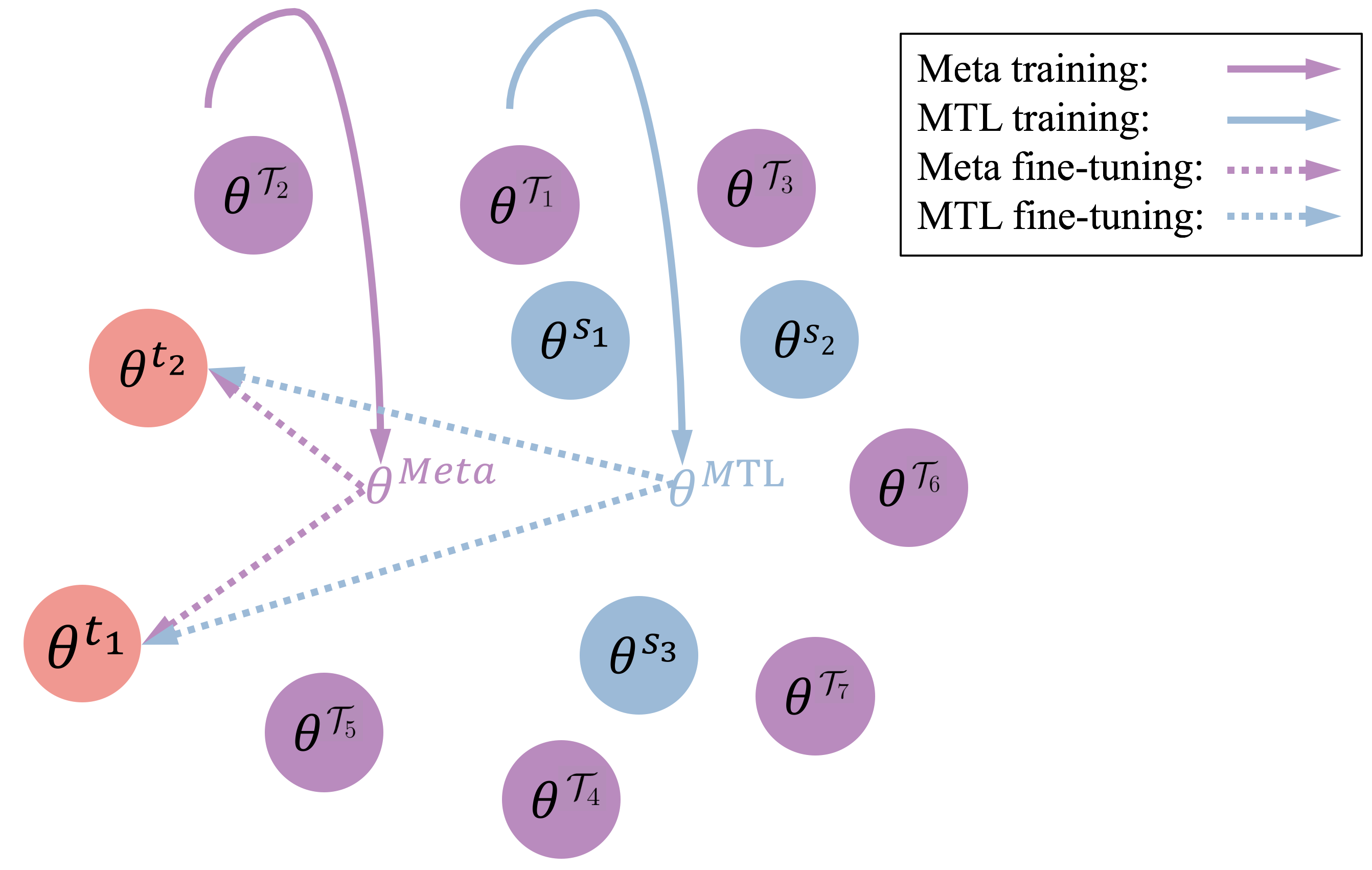}
		\vspace{-0.05in}
		\caption{Comparing Meta-Learning to Multi-task Learning: $\theta^{Meta}$ meta-learned from auxiliary \textit{{Meta NLG tasks}} can be fine-tuned easier than $\theta^{MTL}$ to some new low-resource tasks, e.g, $t_1$ and $t_2$.}
		\label{fig:meta_formulation}
		\vspace{-0.15in}
	\end{figure}
\vspace{-0.01in}
    \begin{equation}
	\small
	\theta^{MTL} = MTL (\mathcal{D}_{\mathbf{s}})= \arg \max_{\theta} \mathbb{E}_{j}  \mathcal{L}_{\mathcal{D}_{s_j}}(f_{\theta})
	\vspace{-0.03in}
	\label{eq:mtl}
	\end{equation}
	
	To better understand the meta objective, we include a standard multi-task learning~(MTL) objective in Eq.(\ref{eq:mtl}).
	MTL learns through individual DA-utterance pairs from different high-resource NLG tasks ${s_j}$, and it does not explicitly learn to adapt to new low-resource NLG tasks.
	Figure~\ref{fig:meta_formulation} visually illustrates the differences with three high-resource source tasks $ \{s_1, s_2, s_3\}$with optimal parameters $\{\theta^{s_1}, \theta^{s_2}, \theta^{s_3}\}$ for each task. $\theta^{MTL}$ is learned from individual DA-utterance pairs in $\{\mathcal{D}_{s_1}, \mathcal{D}_{s_2}, \mathcal{D}_{s_3}\}$, while \meta\ repeatedly constructs auxiliary \textit{{Meta NLG tasks}}  $\{\mathcal{T}_1, ..., \mathcal{T}_7\}$ from $\{\mathcal{D}_{s_1}, \mathcal{D}_{s_2}, \mathcal{D}_{s_3}\}$ and directly learns $\theta^{Meta}$ from them. As a result, $\theta^{Meta}$ is closer to $\theta^{t_1}$ and $\theta^{t_2}$  (the optimal parameters of some new low-resource tasks, e.g, $t_1$ and $t_2$) than $\theta^{MTL}$.
    As we will see soon later, our meta optimization scheme results in a substantial gain in the final performance.
    
	Algorithm 1 illustrates the process to learn  $\theta^{Meta}$ from $\mathcal{D}_\mathbf{s}$. We note that batches are at the level of \textit{Meta NLG tasks}, not DA-utterances pairs. Fine-tuning \meta\ on a new low-resource NLG task with annotated DA-utterance pairs $\mathcal{D}_t$ uses the same algorithm parameterized by ($f_{\theta}, \theta^{\mathbf{s}}, \mathcal{D}_t, \alpha, \beta$). 
	
	\begin{algorithm}
		\fontsize{8.6}{8.6}\selectfont
		\begin{algorithmic}[1]
			
			\REQUIRE
			{$f_{\theta}, \theta_0, \mathcal{D}_\mathbf{s}, \alpha, \beta$}
			\ENSURE  {$\theta^{Meta}$ }
			
			\STATE Initialize $\theta = \theta_0 $
			
			
			\WHILE{$\theta$ not converge} 
			
			\STATE Simulate a batch of \textit{Meta NLG tasks} $\{\mathcal{T}_i = ( \mathcal{D}_{\mathcal{T}_i}, \mathcal{D}_{\mathcal{T}_i}^{'})\}_{i=1}^K$
			
			\FOR {$ i = 1...K$ }
			
			
			
			\STATE Compute $\theta^{'}_i  = \theta - \alpha \nabla_\theta \mathcal{L}_{\mathcal{D}_{\mathcal{T}_i}} (f_{\theta})$ in Eq.(\ref{eq:inner})
			
			
			\ENDFOR
			
			
			\STATE Meta update 
			$\theta = \theta - \beta \sum_{i=1}^{K}\nabla_\theta \mathcal{L}_{\mathcal{D}_{\mathcal{T}_i}^{'}}(f_{\theta_i^{'}})$ in Eq.(\ref{eq:outer})
			
			\ENDWHILE
			
			\STATE Return $\theta^{Meta}  = \theta$
			
			\caption{\small Meta-NLG($f_{\theta}, \theta_0, \mathcal{D}_\mathbf{s}, \alpha, \beta$)}
			
			\label{alg:one}
			
		\end{algorithmic}	
	\end{algorithm}
	\vspace{-0.05in}

	\section{Experiment}

	\subsection{Baselines and Model Settings}
	
	We utilized the well-recognized semantically conditioned LSTM~(SCLSTM \cite{wen2015semantically}) as the base model $f_{\theta}$. 
	We used the default setting of hyperparameters (n\_layer = 1, hidden\_size = 100, dropout = 0.25, clip = 0.5, beam\_width = 5). 
    We implemented \meta\ based on the PyTorch SCLSTM implementation from~\cite{budzianowski2018multiwoz}.
	As \meta\ is model-agnostic, it is applicable to many other NLG models. 
  
	We included different model settings as baseline:

	\begin{itemize}
		
		\item \textbf{\scratch:} Train $f_{\theta}$ with only low-resource target task data, ignoring all high-resource source task data.
		
		\item \textbf{\mtl:} Train $f_{\theta}$ using a multi-task learning paradigm with source task data, then fine-tune on the low-resource target task.		
		
		\item \textbf{\zero:} Train $f_{\theta}$ using multi-task learning (MTL) with source task data, then directly test on a target task without a fine-tuning step. This corresponds to a zero-shot learning scenario.
		
		\item \textbf{\supervise:} Train $f_{\theta}$ using MTL with full access to high-resource data from both source and target tasks. Its performance serves an upper bound using multi-task learning without the low-resource restriction.
		
		\item \textbf{\meta (proposed):} Use Algorithm 1 to train $f_{\theta}$ on source task data, then fine-tune on the low-resource target task. 		
		
	\end{itemize}
	
	For \meta, we set batch size to 5, and $\alpha = 0.1$ and $\beta=0.001$. A single inner gradient update is used per meta update with Adam~\cite{Kingma2014adam}.
	The size of a \textit{Meta NLG task} is set to 400 with 200 samples assigned to ${\mathcal{D}_{\mathcal{T}_i}}$ and ${\mathcal{D}_{\mathcal{T}_i}^{'}}$ because the minimum amount of target low-resource samples is 200 in our later experiments.
	During fine-tuning on a low-resource target task, early-stop is conducted on a small validation set with size 200. The model is then evaluated on other DA-utterance pairs in the target task. 
    
      As in earlier NLG researches, we use the BLEU-4 score~\cite{papineni2002bleu} and the slot error rate (ERR) as evaluation metrics.
 ERR is computed by the ratio of the sum of the number of missing and redundant slots in a generated utterance divided by the total number of slots in the DA.
 	We randomly sampled target low-resource task five times for each experiment and reported the average score.

	\subsection{MultiWoz Dataset for NLG}
\begin{figure}[htb!]
		\centering
		\includegraphics[width=0.46\textwidth]{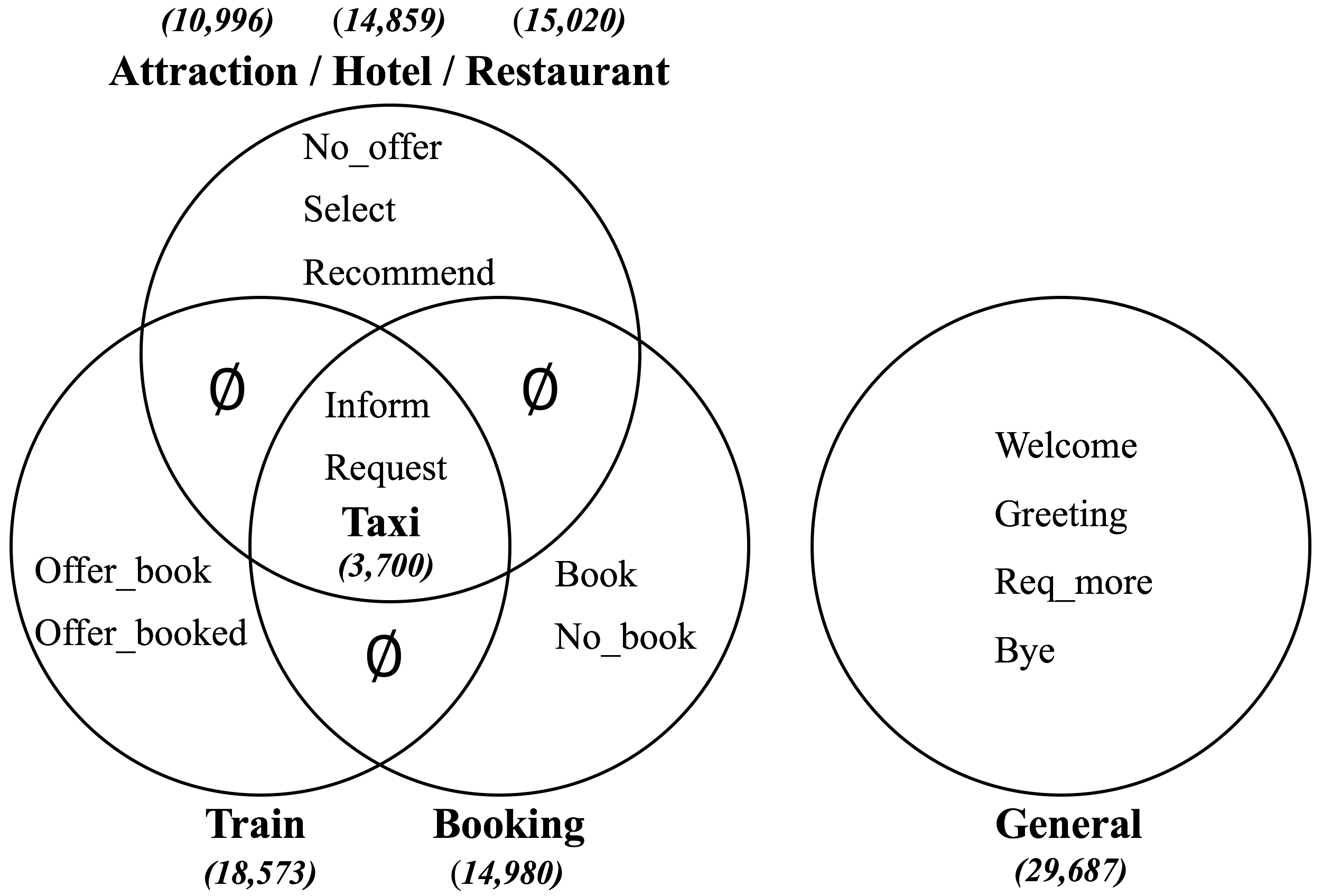}
		\vspace{-0.1in}
		\caption{DA type visualization in different domains. Number of utterances in each domain is indicated in bracket.}
		\label{fig:data}
		\vspace{-0.15in}
	\end{figure}

            We used a recently proposed large-scale multi-domain dialog dataset~(MultiWOZ,~\cite{budzianowski2018multiwoz}). It is a proper benchmark for evaluating NLG components due to its domain complexity and rich linguistic variations.
    A visualization of DA types in different domains are given in Figure~\ref{fig:data}, and slots in different domains are summarized in Table~\ref{table:data}.
	The average utterance length is 15.12, and almost 60\% of utterances have more than one dialogue act  types or domains.
	A total of 69,607 annotated utterances are used, with 55,026, 7,291, 7,290 utterances for training, validation, and testing respectively.
    
 \vspace{-0.05in}
	
	\begin{table}[htb!]
		\centering
		\fontsize{9}{9}\selectfont
        \tabcolsep=0.18cm
		\begin{tabu}{c|c}		
			\hline
			\textbf{Attraction} & Addr, Area ,  Choice,  \textbf{Fee},  Name,  \textbf{Open},  \\
			& Phone,  Post ,  Price,  Type   \\ \hline
			\textbf{Hotel} &   Addr,  Area,  Choice,  \textbf{Internet},  Name,  \textbf{Parking}, \\
			& Phone,  Post,  Price,  Ref,  \textbf{Stars},  Type  \\ \hline
			\textbf{Restaurant} &   Addr,  Area,  Choice,  \textbf{Food},  Name,  Phone,  \\
			& Post,  Price,  Ref  \\ \hline
			\textbf{Train}&   Arrive,  Choice,  Day,  Depart,  Dest,  \textbf{Id}, \\
			& Leave,  People,  Ref,  \textbf{Ticket},  Time  \\  \hline
			\textbf{Booking}&   Day,  Name,  \textbf{People},  Ref,  \textbf{Stay},  Time  \\ \hline
			\textbf{Taxi} &   Arrive,  \textbf{Car},  Depart,  Dest,  Leave,  Phone \\ \hline
			\textbf{General} & None \\ \hline
			
		\end{tabu}
		\vspace{-0.1in}
		\caption{Slots in each domain, with domain-specific slots in bold. }
		\label{table:data}
		\vspace{-0.1in}
	\end{table}

\begin{table*}[htb!]
		\centering
		\fontsize{8.5}{8.5}\selectfont
		\tabcolsep=0.15cm
		\begin{tabu}{l|cr|rr|rr|cr|rr|rr}
			\hline
			\rowfont{\normalsize}	& \multicolumn{6}{c|}{Target Domain = \textbf{Attraction}}                                                                                                                                            & \multicolumn{6}{c}{Target Domain = \textbf{Hotel} }                                                                                                                                                       \\ \cline{2-13} 
			
			& \multicolumn{3}{c|}{\textbf{\supervise}}                                     & \multicolumn{3}{c|}{\textbf{\zero}}                                                                    &                \multicolumn{3}{c|}{\textbf{\supervise}}                                                & \multicolumn{3}{c}{\textbf{\zero}}                                                                                                    \\
			& BLEU-4                               & \multicolumn{2}{c|}{ERR}             & \multicolumn{1}{c}{BLEU-4}                               & \multicolumn{2}{c|}{ERR}                              &                                 BLEU-4                               &  \multicolumn{2}{c|}{ERR}                                   & \multicolumn{1}{c}{BLEU-4 }                               &  \multicolumn{2}{c}{ERR}                                                                      \\
			& 0.5587                               & \multicolumn{2}{c|}{3.05\%  }                           & \multicolumn{1}{c}{0.2970}                               & \multicolumn{2}{c|}{11.56\%}                                                 & 0.4393                               & \multicolumn{2}{c|}{1.82\%}                             & \multicolumn{1}{c}{0.2514}                                & \multicolumn{2}{c}{13.40\% }                                                                           \\ \cline{2-13} 
			& \multicolumn{2}{c|}{Adapt 1000}                                & \multicolumn{2}{c|}{Adapt 500}                        & \multicolumn{2}{c|}{Adapt 200}                        & \multicolumn{2}{c|}{Adapt 1000}                                & \multicolumn{2}{c|}{Adapt 500}                        & \multicolumn{2}{c}{Adapt 200}                                 \\ \cline{2-13} 
			& BLEU-4                              & \multicolumn{1}{c|}{ERR} & \multicolumn{1}{c}{BLEU-4} & \multicolumn{1}{c|}{ERR} & \multicolumn{1}{c}{BLEU-4} & \multicolumn{1}{c|}{ERR} & BLEU-4                              & \multicolumn{1}{c|}{ERR} & \multicolumn{1}{c}{BLEU-4} & \multicolumn{1}{c|}{ERR} & \multicolumn{1}{c}{BLEU-4}    & \multicolumn{1}{c}{ERR}        \\
			
			\multicolumn{1}{c|}{\textbf{\scratch}} & \multicolumn{1}{r}{0.5102}          & 21.84\%                  & 0.4504                     & 36.50\%                  & 0.4089                     & 41.83\%                  & \multicolumn{1}{r}{0.3857}          & 18.75\%                  & 0.3529                     & 28.18\%                  & 0.2910                        & 40.86\%                        \\
			
			\multicolumn{1}{c|}{\textbf{\mtl}} & \multicolumn{1}{r}{0.5443}          & 13.04\%                  & 0.5324                     & 14.34\%                  & 0.4912                     & 23.20\%                  & \multicolumn{1}{r}{0.4128}          & 9.93\%                  & 0.3802                     & 22.07\%                  & 0.3419                        & 31.04\%                        \\
			
			\multicolumn{1}{c|}{\textbf{\meta}}    & \multicolumn{1}{r}{\textbf{0.5667}} & \textbf{2.26\%}          & \textbf{0.5662}            & \textbf{2.97\%}          & \textbf{0.5641}            & \textbf{4.30\%}         & \multicolumn{1}{r}{\textbf{0.4436}} & \textbf{1.92\%}          & \textbf{0.4365}            & \textbf{2.63\%}         & \textbf{0.4418}               & \textbf{2.19\%}   \\
			\hline
			
		\end{tabu}
		\vspace{-0.1in}
		\caption{Results for near-domain adaption to ``Attraction'' and ``Hotel'' domain, with different adaptation sizes.}
		\vspace{-0.06in}
		\label{table:near}
	\end{table*}
	
	\begin{table*}[htb!]
		\centering
		\fontsize{8.5}{8.5}\selectfont
		\tabcolsep=0.15cm
		\begin{tabu}{l|cr|rr|rr|cr|rr|rr}
			\hline
			\rowfont{\normalsize} & \multicolumn{6}{c|}{Target Domain = \textbf{Booking}}                                                                                                                                            & \multicolumn{6}{c}{Target Domain = \textbf{Train} }                                                                                                                                                       \\ \cline{2-13} 
			& \multicolumn{3}{c|}{\textbf{\supervise}}                                     & \multicolumn{3}{c|}{\textbf{\zero}}                                                                    &            \multicolumn{3}{c|}{\textbf{\supervise}}                                             & \multicolumn{3}{|c}{\textbf{\zero}}                                                                                                    \\
			& BLEU-4                               & \multicolumn{2}{c|}{ERR}             & \multicolumn{1}{c}{BLEU-4}                               & \multicolumn{2}{c|}{ERR}                              &                                 BLEU-4                               &  \multicolumn{2}{c|}{ERR}                                   & \multicolumn{1}{c}{BLEU-4 }                               &  \multicolumn{2}{c}{ERR}                                                                      \\
			& 0.6750                               & \multicolumn{2}{c|}{3.67\%  }                           & \multicolumn{1}{c}{0.3578}                               & \multicolumn{2}{c|}{12.55\%}                                                 & 0.6877                               & \multicolumn{2}{c|}{2.96\%}                             & \multicolumn{1}{c}{0.3243}                                & \multicolumn{2}{c}{41.48\% }                                                                           \\ \cline{2-13} 
			& \multicolumn{2}{c|}{Adapt 1000}                                & \multicolumn{2}{c|}{Adapt 500}                        & \multicolumn{2}{c|}{Adapt 200}                        & \multicolumn{2}{c|}{Adapt 1000}                                & \multicolumn{2}{c|}{Adapt 500}                        & \multicolumn{2}{c}{Adapt 200}                                 \\ \cline{2-13} 
			& BLEU-4                              & \multicolumn{1}{c|}{ERR} & \multicolumn{1}{c}{BLEU-4} & \multicolumn{1}{c|}{ERR} & \multicolumn{1}{c}{BLEU-4} & \multicolumn{1}{c|}{ERR} & BLEU-4                              & \multicolumn{1}{c|}{ERR} & \multicolumn{1}{c}{BLEU-4} & \multicolumn{1}{c|}{ERR} & \multicolumn{1}{c}{BLEU-4}    & \multicolumn{1}{c}{ERR}        \\
			
			\multicolumn{1}{c|}{\textbf{\scratch}} & \multicolumn{1}{r}{0.6327}          & 24.63\%                  & 0.6267                     & 37.96\%                  & 0.5787                     & 46.67\%                  & \multicolumn{1}{r}{0.6236}          & 16.73\%                  & 0.5825                     & 27.61\%                  & 0.4892                        & 44.92\%                        \\
			
			\multicolumn{1}{c|}{\textbf{\mtl}}     & \multicolumn{1}{r}{0.6347}          & 14.55\%                  & 0.6391                     & 14.90\%                  & 0.6171                     & 17.19\%                  & \multicolumn{1}{r}{0.6322}          & 14.63\%                  & 0.5987                     & 25.38\%                  & {\color[HTML]{000000} 0.5248} & {\color[HTML]{000000} 40.35\%} \\
			
			\multicolumn{1}{c|}{\textbf{\meta}}    & \multicolumn{1}{r}{\textbf{0.6782}} & \textbf{7.65\%}          & \textbf{0.6492}            & \textbf{9.08\%}          & \textbf{0.6402}            & \textbf{12.23\%}         & \multicolumn{1}{r}{\textbf{0.6755}} & \textbf{7.13\%}          & \textbf{0.6373}            & \textbf{17.31\%}         & \textbf{0.6160}               & \textbf{23.33\%}     \\      
			\hline
			
		\end{tabu}
		\vspace{-0.1in}
		\caption{Results for far-domain adaption to ``Booking'' and ``Train'' domain, with different adaptation sizes.}
		\vspace{-0.16in}
		\label{table:far}
	\end{table*} 

	\subsection{Domain Adaptation}
	\vspace{-0.02in}
	In this section, we tested when a NLG model is adapted to two types (\textit{near} and \textit{far}) of low-resource language domains.
	Experiment follows a \textit{leave-one-out} setup by leaving one target domain for low-resource adaptation, while using the remainder domains as high-resource source training data.
	A target domain is a \emph{near-domain} if it contains no domain-specific DA type but only domain-specific slots compared to the remainder domains. In contrast, a target domain containing both domain-specific DA types and slots is considered as a \emph{far-domain}.
	Adapting to near-domains requires to capture unseen slots, while adapting to far-domains requires to learn new slots as well as new language patterns.
	\textit{Adaptation size} is the number of DA-utterance pairs in the target domain used to fine-tune the NLG model. To test different low-resource degrees, we considered different \textit{adaptation sizes}~(1,000, 500, 200) in subsequent experiments.
	
    
	\textbf{Near-domain Adaptation:} Figure~\ref{fig:data} and Table~\ref{table:data} show that ``Attraction'', ``Hotel'', ``Restaurant'', and ``Taxi'', are four near-domains compared to remainder domains. 
    Only results for ``Attraction'' and ``Hotel'' are included due to page limit. The other two domains are  also simpler with only one domain-specific slot. 
	Several observations can be noted from results in Table \ref{table:near}.
	\textbf{First}, Using only source or target domain samples does not produce competitive performance.  Using only source domain samples (\zero) performs the worst. It obtains very low BLEU-4 scores, indicating that the sentences generated do not match the linguistic patterns in the target domain. Using only low-resource target domain samples (\scratch) performs slightly better, yet still much worse than \mtl\ and \meta.
	\textbf{Second}, \meta\ shows a very strong performance for this near-domain adaptation setting. It consistently outperforms \mtl\ and other methods with very remarkable margins in different metrics and adaptation sizes. More importantly, it even works better than \supervise\ which is trained on high-resource samples in the target domain. 
	\textbf{Third}, \meta\ is particularly strong in performance when the adaptation size is small. 
	As the adaptation size decreases from 1,000 to 200, the performance of \scratch\ and \mtl\ drops quickly, while \meta\ performs stably well. Both BLEU-4 and ERR even increase in ``Hotel'' domain when the adaptation size decreases from 500 to 200.

	\textbf{Far-domain Adaptation:} In this experiment, we tested the performance when adapting to two low-resource far-domains (``Booking'' and ``Train'').
	Again, we can see that \meta\ shows very strong performance on both far-domains with different adaptation sizes. Similar observations can be made as in the previous near-domain adaptation experiments.
	Because far-domain adaptation is more challenging, \meta\ does not outperform \supervise, and the performance of \meta\ drops more obviously as the adaptation size decreases.
	Noticeably, ``Train'' is more difficult than ``Booking'' as the former contains more slots, some of which can only be inferred from the smallest ``Taxi'' domain. The improvement margin of \meta\ over \mtl\ and other methods is larger on the more difficult ``Train'' domain than on the ``Booking'' domain. 
	

	\subsection{Dialog Act (DA) Type Adaptation}
\vspace{-0.1in}
	\begin{table}[htb!]
		\centering
		\fontsize{8.5}{8.5}\selectfont
		\tabcolsep=0.25cm
		\begin{tabu}{l|cr|cr}
			\hline
			\rowfont{\normalsize} & \multicolumn{2}{c|}{\textbf{Book}}     & \multicolumn{2}{c}{ \textbf{Recommend} }    \\
			& BLEU-4 & ERR & BLEU-4 & ERR \\
			\hline
			\multicolumn{1}{c|}{\textbf{\scratch}} & 0.7689          & 21.63\%                  & 0.3878                     & 24.62\%           \\
			
			\multicolumn{1}{c|}{\textbf{\mtl}}     & 0.7968         & 9.92\%                  & 0.3964                     & 14.60\%        \\
			
			\multicolumn{1}{c|}{\textbf{\meta}}    & \textbf{0.8217} & \textbf{4.65}\%          & \textbf{0.4445}            & \textbf{3.08}\%       \\      
			\hline
		\end{tabu}
		\vspace{-0.1in}
		\caption{Results for adapting to new DA type ``Book'' and ``Recommend'' with adaptation size 500.} 
		\vspace{-0.1in}
		\label{table:function}
	\end{table}
	
	\begin{table*}[htb!]
		\centering
		\fontsize{8.8}{8.}\selectfont
		\tabcolsep=0.23cm
		\begin{tabu}{r|l}
			\hline
			\multicolumn{2}{c}{{Inform (\textbf{Ticket}$^{\dag}$=\textbf{17.60 pounds}, \textbf{Time}=\textbf{79 minutes}); Offer\_Book$^\star$(None) }} \\ \hline
			\textbf{Reference} & \textit{The travel time is \textbf{[79 minutes]} and the cost is\textbf{ [17.60 pounds]}, shall I book for you?}  \\ \hline
			\textbf{\mtl} & \textit{there is a train that leaves at \textbf{[slot-train-leave]} and arrives at \textbf{[slot-train-arrive]}. would you like me to book it for you?} \\& \blue{[\textbf{missed}: \textbf{Ticket}$^{\dag}$, \textbf{Time};} \textbf{}\orange{\textbf{redundant}: \textbf{Leave}, \textbf{Arrive}]}\\ \hline
			\textbf{\meta} & \textit{the travel time is \textbf{[79 minutes]} and the price is \textbf{[17.60 pounds]}. would you like me to book it for you?} \red{[\textbf{correct}]}\\\hline
			
			\multicolumn{2}{c}{{Inform(\textbf{Arrive}=\textbf{7:52}, \textbf{Id}$^{\dag}$=\textbf{TR9641}, \textbf{Dest}$^{\ddag}$=\textbf{cambridge}, \textbf{Depart}$^{\ddag}$=\textbf{the airport}, \textbf{Leave} =\textbf{7:24})}}  \\ \hline
			\textbf{Reference} & \textit{\textbf{[TR9641]} leaves \textbf{[the airport]} at \textbf{[7:24]} and arrives in \textbf{[cambridge]} at \textbf{[7:52]}.} \\ \hline
			\textbf{\mtl} & \textit{i have a train that leaves \textbf{[the airport]} at \textbf{[7:24]} and arrives by\textbf{ [7:52]}.} \blue{[\textbf{missed}: \textbf{Id}$^{\dag}$, \textbf{Dest}$^{\ddag}$]} \\ \hline
			\textbf{\meta} & \textit{\textbf{[TR9641]} leaves \textbf{[the airport]} at \textbf{[7:24]} and arrives in \textbf{[cambridge]} at \textbf{[7:52]}.} \red{[\textbf{correct}]} \\ \hline
			
			\multicolumn{2}{c}{{Offer\_Booked$^\star$}(\textbf{Ref}=\textbf{Y8XKP9LU}, \textbf{People}=\textbf{6}, \textbf{Ticket}$^{\dag}$=\textbf{85.92 GBP}, \textbf{Id}$^{\dag}$=\textbf{TR4967}, \textbf{Depart}$^{\ddag}$=\textbf{Broxbourne}, \textbf{Leave}=\textbf{19:32}); Req\_more(None)}  \\ \hline
			\textbf{Reference} & \textit{I booked \textbf{[6]} tickets for you on \textbf{[TR4967]} leaving \textbf{[Broxbourne]} at \textbf{[19:32]}. \textbf{ [85.92 GBP]} is payable at the station.  }\\ & \textit{Reference number is \textbf{[Y8XKP9LU]}.  Will that be all? }\\ \hline
			
			\textbf{\mtl} & \textit{i have booked a train that leaves at \textbf{[19:32]}. your reference number is \textbf{[Y8XKP9LU]}.} \\ &  \textit{is there anything else i can help you with? }\blue{[\textbf{missed}: \textbf{People}, \textbf{Ticket}$^{\dag}$, \textbf{Id}$^{\dag}$, \textbf{Depart}$^{\ddag}$]} \\ \hline
			
			\textbf{\meta} & \textit{i was able to book a train \textbf{[TR4967]} leave \textbf{[Broxbourne]} at \textbf{[19:32]}. the total fee is \textbf{[85.92 GBP]} payable at the station.} \\ & \textit{your reference number is \textbf{[Y8XKP9LU]}. is there anything else i can help you with?} \blue{[\textbf{missed}: \textbf{People}]} \\ \hline
		\end{tabu}
		\vspace{-0.1in}
		\caption{Sampled generated sentences when considering ``Train'' as the target domain with  adaptation size 500. Slots that are missed or redundant are colored in blue and orange respectively. 
			$^\star$ indicates a \textit{domain-specific DA type}, $^{\dag}$ indicates a \textit{domain-specific slot}, and $^{\ddag}$ indicates a\textit{ rare slot} that can only be inferred from the smallest ``Taxi'' domain.}
		\label{table:samples}
		\vspace{-0.15in}
	\end{table*}
	
	It is also important and attractive for a task-oriented dialog system to adapt to new functions, namely, supporting new dialog acts that the system has never observed before. 
	To test this ability, we left certain DA types out for adaptation in a low-resource setting. We chose ``Recommend'', ``Book'' as target DA types, and we mimic the situation that a dialog system needs to add a new function to make \textit{recommendations} or \textit{bookings} for customers with a few number of annotated DA-utterance pairs.
	As presented in Table~\ref{table:function}, results show that \meta\ significantly outperforms other baselines. 
	Therefore, we can see that \meta\ is also able to adapt well to new functions that a dialog system has never observed before.
	
	\subsection{Adaptation Curve Analysis}
\vspace{-0.12in}
	\begin{figure}[htb!]
		\centering
		\includegraphics[width=0.48\textwidth]{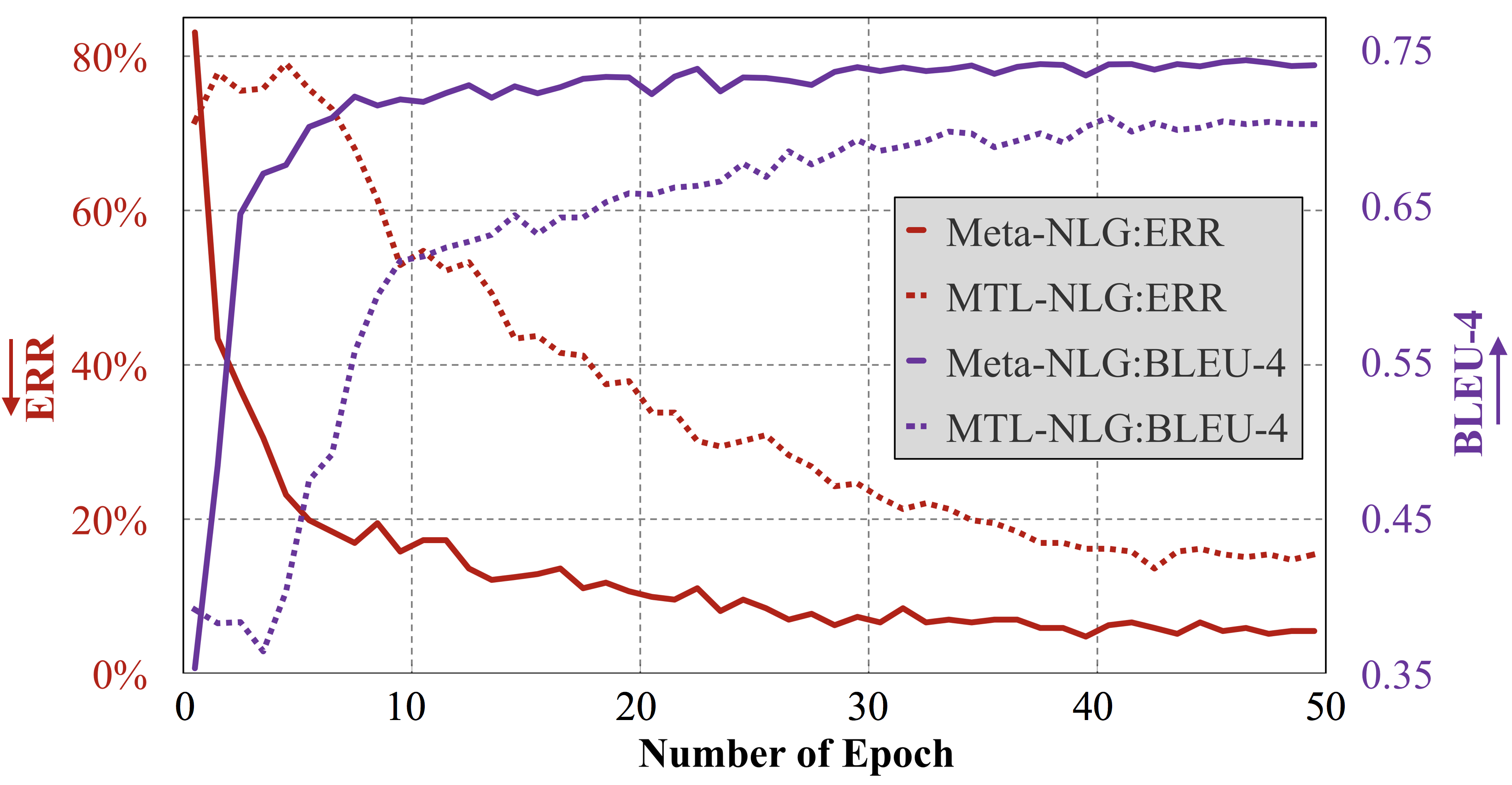}
		\vspace{-0.2in}
		\caption{\small ERRs (red) and BLEU-4 (purple) scores of \meta\ and \mtl\ on the validation set during model fine-tuning on the target low-resource domain~(Train) with adaptation size 500.}
		\label{fig:curve}
		\vspace{-0.12in}
	\end{figure}

	To further investigate the adaptation process, we presented  in Figure~\ref{fig:curve} the performance curves of \mtl\ and \meta\ as fine-tuning epoch proceeds on the most challenging ``Train'' domain. 
	The effect of meta-learning for low-resource NLG can be observed by comparing the two solid curves against the corresponding dashed curves.
	\textbf{First}, \meta\ adapts \textit{faster} than \mtl. We can see that the ERR of \meta\ (red-solid) decreases much more rapidly than that of \mtl\ (red-dashed) , and the BLEU-4 score of \meta\ (purple-solid)  also increases more quickly. The optimal BLEU-4 and ERR that \mtl\ converges to can be obtained by \meta\ within 10 epochs. 
	\textbf{Second}, \meta\ adapts \textit{better} than \mtl. As it can be seen, \meta\ achieves a much lower ERR and a higher BLEU-4 score when it converges, indicating that it found a better $\theta$ of the base NLG model to generalize to the low-resource target domain.


	\subsection{Manual Evaluation}
	To better evaluate the quality of the generated utterances, we performed
	manual evaluation. 
    
    \textbf{Metrics:} Given a DA and a reference utterance in a low-resource target domain with adaptation size 500, two
	responses generated by \meta\ and \mtl\ were presented to three human annotators to score each of them in terms of \textit{informativeness} and \textit{naturalness} (rating out of 3), and also indicate their \textit{pairwise preferences} (Win-Tie-Lose) on \meta\ against \mtl.
    \textit{Informativeness} is defined as whether the generated utterance captures all the information, including multiple slots and probably multiple DA types, specified in the DA.
    \textit{Naturalness} measures whether the utterance is plausibly generated by a human. 

    \textbf{Annotation Statistics:} Cases with identical utterances generated by two models were filtered out.
	We obtained in total 600 annotations on each individual metric for each target domain. We calculated the Fleiss’ kappa \cite{fleiss1971measuring} to measure inter-rater consistency. The overall Fleiss’ kappa values for {\textit{informativeness}} and {\textit{naturalness}} are 0.475 and 0.562, indicating ``Moderate Agreement'', and 0.637 for \textit{{pairwise preferences}}, indicating “Substantial Agreement”.
    
   \textbf{Results:} Scores of \textit{informativeness} and \textit{naturalness} are presented in Table~\ref{table:human1}. \meta\ outscores \mtl\ in terms of both metrics on all four domains. Overall, \meta\ received significantly (two-tailed  t-test, $p<0.0005$) higher scores than \mtl.  
   Results for \textit{pairwise preferences} are summarized in Table~\ref{table:human2}. Even though there are certain amount of cases where the utterances generated by different models are nearly the same (Tie) to annotators, \meta\ is overall \textit{significantly} preferred over \mtl ~(two-tailed  t-test, $p<0.0001$) across different target domains.  
  
  \vspace{-0.05in}
      	\begin{table}[htb!]
		\centering
		\fontsize{8.7}{8.7}\selectfont
        \tabcolsep=0.07cm
		\begin{tabu}{c|cc|cc|cc|cc|cc}
			\hline
			&  \multicolumn{2}{c|}{\textbf{Overall}}  & \multicolumn{2}{c|}{\textbf{Attraction}}   & \multicolumn{2}{c|}{\textbf{Hotel}}  &\multicolumn{2}{c|}{\textbf{Booking}}  &\multicolumn{2}{c}{\textbf{Train}} \\
			\cline{2-11} 
            & inf. & nat. & inf. & nat. & inf. & nat. & inf. & nat. & inf. & nat.  \\
            \hline       
             \meta\ & 2.85 & 2.91 & 2.91 & 2.90 & 2.90 & 2.89 & 2.84 & 2.91 & 2.73 & 2.93 \\
             \mtl\ & 2.60 & 2.85 & 2.70 & 2.87 & 2.57 & 2.83 & 2.65 & 2.86 & 2.47 & 2.83 \\
			\hline
		\end{tabu}
		\vspace{-0.1in}
		\caption{Manual evaluation scores (rating out of 3) with \textit{informativeness} (inf.), and \textit{naturalness} (nat.) on target low-resource domains. The overall scores (column 2) are aggregated from four domains.}
		\vspace{-0.15in}
		\label{table:human1}
	\end{table}
    \vspace{-0.07in}
    \begin{table}[htb!]
		\centering
		\fontsize{8.7}{8.7}\selectfont
		\begin{tabu}{c|c|c|c|c|c}
			\hline
			&  \textbf{Overall} &  \textbf{Attraction}  & \textbf{Hotel}  & \textbf{Booking}  & \textbf{Train}  \\
			\hline
			Win & 47.7\% & {50.2}\% & {53.3}\% & 40.1\% & {47.2}\% \\
			Tie & 42.9\% & 42.8\% & 42.3\% & {46.2}\% & 40.5\% \\
			Lose & 9.4\% & 7.0\% & 4.4\% & 13.7\% &12.3\%\\
			\hline
		\end{tabu}
		\vspace{-0.1in}
		\caption{\textit{Pairwise preferences} (\meta\ vs. \mtl) on target low-resource domains. The overall preferences (column 2) are aggregated from four domains.}
		\vspace{-0.15in}
		\label{table:human2}
	\end{table}
	
	\subsection{Case Study}
	
	Table~\ref{table:samples} shows three examples in the ``Train'' domain.
	The first sample shows that \mtl\ fails to generate the domain-specific slot ``Ticket'', instead, it mistakenly generates slots (``Leave'' and ``Arrive'') that are frequently observed in the low-resource adaptation set.
	In the second example, \mtl\ failed to generate the domain-specific slot `Id'' and another rare slot ``Dest'', while \meta\ succeeded both.
	The last example shows similar results on a domain-specific dialog act type ``Offer\_Booked'', in which \meta\ successfully captured two domain-specific slots and a rare slot.

	\vspace{-0.05in}
	\section{Conclusion}
	
	We propose a generalized optimization-based meta-learning approach \meta\ for the low-resource NLG task.
	\meta\ utilizes \textit{Meta NLG tasks} and a meta-learning optimization procedure based on MAML.
	Extensive experiments on a new benchmark dataset (MultiWoz) show that \meta\ significantly outperforms other training procedures, indicating that our method adapts fast and well to new low-resource settings. Our work may inspire researchers to use similar optimization techniques for building more robust and scalable NLG components in task-oriented dialog systems.
	
    \small
	\bibliographystyle{named}
	\bibliography{meta}
	
\end{document}